\pdfoutput=1
\documentclass[runningheads]{llncs}
\usepackage{graphicx}
\usepackage{comment}
\usepackage{amsmath,amssymb} 
\usepackage{epsfig}
\usepackage{caption}

\usepackage{color}
\usepackage[caption=false]{subfig}
\usepackage{float}
\usepackage[dvipsnames]{xcolor}
\definecolor{myorange}{rgb}{1.0,0.549,0.0}

\begin{document}
\pagestyle{headings}
\mainmatter
\def\ECCVSubNumber{2540}  

\title{Topology-Change-Aware Volumetric Fusion for Dynamic Scene Reconstruction} 

\titlerunning{Topology-change-aware Fusion}
%
\author{Chao Li \and Xiaohu Guo}
\authorrunning{C. Li and X. Guo}
%
\institute{Department of Computer Science,\\
	The University of Texas at Dallas\\
	\email{ \{Chao.Li2, xguo\}@utdallas.edu}
	}
\maketitle

\begin{abstract}
Topology change is a challenging problem for 4D reconstruction of dynamic scenes. In the classic volumetric fusion-based framework, a mesh is usually extracted from the TSDF volume as the canonical surface representation to help estimating deformation field. However, the surface and Embedded Deformation Graph (EDG) representations bring conflicts under topology changes since the surface mesh has fixed-connectivity but the deformation field can be discontinuous. 
In this paper, the classic framework is re-designed to enable 4D reconstruction of dynamic scene under topology changes, by introducing a novel structure of Non-manifold Volumetric Grid to the re-design of both TSDF and EDG, which allows connectivity updates by cell splitting and replication.
Experiments show convincing reconstruction results for dynamic scenes of topology changes, as compared to the state-of-the-art methods.
\keywords{Reconstruction, Topology Change, Fusion, Dynamic Scene}
\end{abstract}

\section{Introduction}
As the development of Virtual Reality, Augmented Reality and 5G technologies, the demand on 4D reconstruction  (space + time) techniques has been raised. 
Especially with the latest advancements of consumer-level RGB-D cameras, the interest has been growing in developing such 4D reconstruction techniques to capture various dynamic scenes. Volumetric fusion-based techniques~\cite{newcombe2015dynamicfusion,dou2017motion2fusion,yu2018doublefusion} allow the 4D reconstruction of dynamic scenes with a single RGB-D camera, by incrementally fusing the captured depth into a volume encoded by Truncated Signed Distance Fields (TSDF)~\cite{Curless1996}.
The philosophy of such volumetric fusion-based reconstruction is to decompose the 4D information into representations of 3D-space and 1D-time individually. The 3D-space information includes two parts: the geometry of the scene is represented in a canonical volume~\cite{newcombe2015dynamicfusion} (or key volumes~\cite{dou2016fusion4d}) encoded by TSDF; the deformation field of the scene is represented by the transformations on an Embedded Deformation Graph (EDG)~\cite{Sumner2007}. 
Along the 1D-time, the deformation field varies and the geometry becomes more complete by fusing more coming frames. 
In order to estimate the deformation field, an intermediate geometry representation, usually a surface mesh, is extracted to solve the model-to-frame registration.
However, in the current fusion framework, this intermediate geometry representation and the EDG built on top of it cannot handle topology change cases when the deformation is discontinuous over 3D space, because they have fixed connectivity between vertices or nodes.

Defining a more flexible data structure to handle topology changes is nontrivial. 
In this paper, our key contribution is the fundamental re-design of the volumetric fusion framework, by revisiting the data structures of geometry and deformation field.
We introduce \emph{Non-manifold Volumetric Grids} into the TSDF representation, by allowing the volumetric grids to replicate themselves and break connections, and design the EDG in a similar non-manifold structure. 
Such a novel design overcomes the issue brought by fixed connectivity of EDG and intermediate mesh (extracted from TSDF grids) and allows their flexible connectivity update throughout the scanning process.

Our second contribution is the proposal of a novel topology-change-aware non-rigid registration method inspired by line process~\cite{black1996}.
This approach efficiently and effectively solves the discontinuity issue due to topology changes by adapting weights to loosen the regularization constraints on edges where topology changes happen.
Based on such a registration framework, we also propose a topology change event detection approach to guide the connectivity updates of EDG and volumetric grids by fully utilizing line process weights.

\section{Related Work}
The most popular methods to reconstruct 4D dynamic scene are using a pre-defined template, such as skeleton~\cite{yu2017bodyfusion}, human body model~\cite{yu2018doublefusion} or pre-scanned geometry~\cite{zollhofer2014real} as prior knowledge, and reconstruct human body parts~\cite{li2013realtime,tkach2016sphere,pons2010multisensor,yu2018doublefusion}.
To eliminate the dependency on such priors, some template-less fusion-based methods were proposed to utilize more advanced structure to merge and store geometry information across motion sequences~\cite{collet2015high,newcombe2015dynamicfusion,innmann2016volumedeform,dou2016fusion4d,guo2017real,dou2017motion2fusion,li2018articulatedfusion,gao2019surfelwarp}. 

However, there are still two major problems related to 4D dynamic scene reconstruction. 
Firstly, all of exiting methods are still vulnerable to fast and occluded motion of dynamic scene. 
Fast motions introduce motion-blur and can severely degrade the tracking accuracy of correspondences between frames which affects geometry fusion. 
The problem is partially solved in \cite{dou2017motion2fusion} by Spectral Embedding, and by \cite{kowdle2018need} with their high frame rate RGB-D sensors. 
The second issue is notorious topology change handling problem, which is our focus here.
Only a few methods are proposed to handle topology changes. 
Key volumes were proposed in \cite{dou2016fusion4d} and \cite{dou2017motion2fusion} to set a new key frame and reinitialize model tracking when a topology change happens.
\cite{slavcheva2017killingfusion} and \cite{slavcheva2018sobolevfusion} propose new methods to tackle this issue by aligning TSDF volumes between two frames. 
However, the resolution of TSDF volume in these methods are lower than that of other mesh-based fusion methods because their fully volumetric registration has scalability limitations. Furthermore, they cannot provide the segmentation information that we offer in our method: separated objects will be reconstructed as independent meshes.

Currently most of the template-less dynamic 4D reconstruction methods use TSDF as the underlying surface representation~\cite{newcombe2015dynamicfusion,innmann2016volumedeform,dou2016fusion4d,yu2017bodyfusion,guo2017real,dou2017motion2fusion,li2018articulatedfusion,oswald2014generalized}. 
However, in dynamic scene reconstruction, the deformation field could be discontinuous, which cannot be represented with a fixed connectivity intermediate mesh and EDG. 
The approach we propose here will allow the dynamic updates to the TSDF volumetric grids conforming to the discontinuity of the deformation fields.
Compared to level set variants \cite{osher2005level,enright2002animation}, which support surface splitting and merging in physical simulation and usually have a noise-free complete mesh, our method aims to incrementally reconstruct geometry from noisy partial scans.

Zampogiannis \textit{et al.}~\cite{zampogiannis2019} proposed a topology-change-aware non-rigid point cloud registration approach by detecting topology change regions based on stretch and compression measurement in both forward and backward motion.
However, how to recover the geometry of dynamic scenes under such topology changes is not explored. 
Inspired by methods in computer animation -- virtual node~\cite{molino2004virtual} and non-manifold level set~\cite{mitchell2015non}, we re-design the non-manifold level set and adapt it to the fusion-based 4D reconstruction framework. Tsoli and Argyros~\cite{tsoli2016tracking} presented a method to track topologically changed deformable surfaces with a pre-defined template given input RGB-D images. 
Compared to their work, our method is template-less, gradually reconstructing the geometry and updating the connectivity of EDG and TSDF volume grids. 
Bojsen-Hansen \textit{et al.}~\cite{bojsen2012tracking} explored in another direction to solve surface tracking with evolving topology. 
But our method can detect topology changes in live frames, recover the changed geometry in the canonical space and playback the entire motion sequence on top of the geometry with new topology. 

There is also a set of works related to dynamic scene reconstruction but not focused on voxel-based techniques: 1) Other template/mesh-based deformation approaches \cite{xu2006poisson,letouzey2012progressive,chen2016mesh}; 2) Methods for learning-based schemes that may handle larger changes \cite{baran2009semantic,frohlich2011example,letouzey2012progressive,gao2013data,von2015real,gao2017data}; 3) Methods on point correspondence based interpolation that do not require the prior of a mesh representation and are more flexible with respect to topological changes \cite{li2012temporally,xu2015deformable,yuan2016space,bertholet2018temporally}; 4) Finally, some point distribution based approaches that do not require correspondence search and provide even more flexibility \cite{digne2014feature,solomon2015convolutional,golla2020temporal}.

\section{System Overview}
\begin{figure*}[htb]
	\centering
	\includegraphics[width=0.95\textwidth]{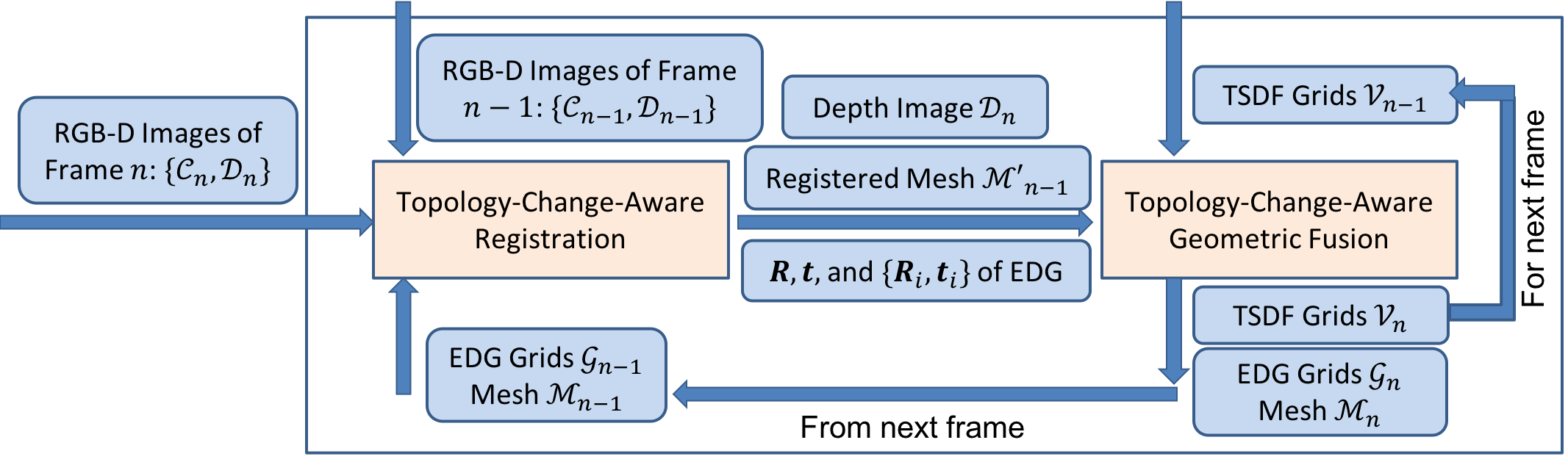}
	\caption{Computational flowchart of our proposed 4D reconstruction system.}
	\label{fig:flowchart}
\end{figure*}

The system takes RGB-D images $\{\mathcal{C}_{n}, \mathcal{D}_{n}\}$ of the $n^{th}$ frame, and outputs a reconstructed surface mesh $\overline{\mathcal{M}_{n}}$ in the canonical space and a per-frame deformation field that transforms that surface into the live frame. 
The topology changes will be reflected by updating the connectivity of EDG and TSDF volume in the canonical space.
In this way, although the topology of $\{\overline{\mathcal{M}_{1}},\cdots,\overline{\mathcal{M}_{n}}\}$ might evolve over time, we can still replicate the topology of the ending frame $\overline{\mathcal{M}_{n}}$ to all of the earlier frames.
Thus we can enable the playback of motions on top of reconstructed meshes with new topology. 
Fig.~\ref{fig:flowchart} shows a flowchart of our 4D reconstruction system, composed of two modules: Topology-Change-Aware Registration, and Topology-Change-Aware Geometric Fusion. 

\section{Technical Details}
Now we describe our reconstruction system
in detail. 
In the first module,
the line process based deformation estimation and non-manifold grid based re-design of EDG are the enabler of topology-change-aware registration.

\subsection{Topology-Change-Aware Registration}
We represent the deformation field through an EDG, of which each node $g^{\mathcal{G}}$ provides a 3DOF displacement $\mathbf{t}_{i}$ for deformation.
For each point (surface vertex or voxel) $\mathbf{x}_{c}$ in canonical space, $\mathbf{T}(\mathbf{x}_{c}) = \mathbf{R}\sum_{i}{\alpha_{i}(\mathbf{x}_{c}+\mathbf{t}_{i})} + \mathbf{t}$ transforms this point from
canonical space into the live frame via trilinear interpolation, where $i$ is the node index of $\mathbf{x}_{c}$-belonged EDG cell and $\alpha_{i}$ is the interpolation weight.
When a new $n^{th}$ frame comes in, we update global rotation $\mathbf{R}$, global translation $\mathbf{t}$, and local displacement $\mathbf{t}_{i}$ on nodes, based on the reconstructed mesh $\mathcal{M}_{n-1}$ from previous frame.

\subsubsection{Estimating The Deformation Field}
The registration can be decomposed into two steps: rigid alignment, and non-rigid alignment. 
The rigid alignment is to estimate the global rotation $\mathbf{R}$ and global translation $\mathbf{t}$ by using dense projective ICP~\cite{rusinkiewicz2001efficient}.
During the non-rigid alignment, we estimate current local deformation field $\{\mathbf{R}_{i},\mathbf{t}_{i}\}$ given the previous reconstructed mesh $\mathcal{M}_{n-1}$ and the RGB-D images $\{\mathcal{C}_{n}, \mathcal{D}_{n}\}$ of this frame by minimizing an energy function.

Similar to VolumeDeform~\cite{innmann2016volumedeform}, we design the energy function as a combination of the following three terms:
\small
\begin{equation}
    E_{total}(\mathbf{X}) = \omega_{s}E_{spr}(\mathbf{X})+\omega_{d}E_{dense}(\mathbf{X})+\omega_{r}E_{reg}(\mathbf{X}),
    \label{eq:total_energy}
\end{equation}
\begin{equation}
    E_{spr}(\mathbf{X}) = \sum_{\mathbf{f}\in \mathcal{F}}\|(\mathbf{T}(\mathbf{f})-\mathbf{y})\|^{2},
    \label{eq:sparse_term}
\end{equation}
\begin{equation}
    E_{dense}(\mathbf{X}) = \sum_{\mathbf{x}\in \mathcal{M}_{n-1}}[\mathbf{n}^{\top}_{y}(\mathbf{T}(\mathbf{x})-\mathbf{y})]^{2}.
    \label{eq:dense_term}
\end{equation}
\normalsize
Here $E_{spr}$ is a sparse feature based alignment term. $E_{dense}$ is a dense depth based measurement and $E_{reg}$ is a regularization term. 
The weights $\omega_{s}, \omega_{d}$ and $\omega_{r}$ control the relative influence of different energy terms.
$\mathbf{y}$ is the corresponding point (in the target) of a feature point or mesh vertex and $\mathbf{n}_{y}$ is the estimated normal of each corresponding point.
We extract the corresponding SIFT features $\mathcal{F}$ between the RGB-D images of current and previous frame as the sparse feature points similar to VolumeDeform~\cite{innmann2016volumedeform}. The dense objective enforces the alignment of the surface mesh $\mathcal{M}_{n-1}$ with the captured depth data based on a point-to-plane distance metric. 
The regularization is an as-rigid-as-possible (ARAP) prior by enforcing the one-ring neighborhood of a node to have similar transformations.
However, such ARAP prior is not able to detect potential topology changes, i.e., the breaking of connection between neighboring nodes.
In this paper, we propose to use a line process~\cite{black1996} to account for the discontinuity caused by topology changes.
The regularization term is:
\small
\begin{equation}
\label{eq:reg_term}
\begin{aligned}
    E_{reg}
    &= \sum_{i}\sum_{j\in{\mathcal{N}(i)}}[l_{ij}\|\mathbf{R}_{i}(\mathbf{g}_{i}-\mathbf{g}_{j})-
    (\Tilde{\mathbf{g}_{i}}-\Tilde{\mathbf{g}_{j}})\|^{2}+\Psi(l_{ij})],
\end{aligned}
\end{equation}

\begin{equation}
\label{eq:g_deform}
     \Tilde{\mathbf{g}_{i}} = \mathbf{g}_{i}+\mathbf{t}_{i},
     \Psi(l_{ij})=\mu(\sqrt{l_{ij}}-1)^{2},
\end{equation}
\normalsize
where $\mathbf{g}_{i}$ and $\mathbf{g}_{j}$ are the positions of the two nodes in EDG $\mathcal{G}_{n-1}$ from previous frame.
The first term in $E_{reg}$ is exactly the ARAP prior measuring the similarity of transformations between neighboring nodes, except for the multiplication of a line process parameter $l_{ij}$ indicating the presence $(l_{ij} \rightarrow 0)$ or absence $(l_{ij} \rightarrow 1)$ of a discontinuity between nodes $i$ and $j$.
The function $\Psi(l_{ij})$ is the ``penalty'' of introducing a discontinuity between the two nodes.
$\mu$ is a weight controlling the balance of these two terms. 
The original strategy of how to set $\mu$ is discussed in paper \cite{black1996}. We will introduce our settings of $\mu$ in detail in next part.
All unknowns to be solved in the entire energy function are:
\begin{equation}
    \mathbf{X} = {(\underbrace{\cdots,\mathbf{R}_{i}^{\top}, \cdots}_\text{rotation matrices} 
    | \underbrace{\cdots,\mathbf{t}_{i}^{\top}, \cdots}_\text{displacements}
    | \underbrace{\cdots,l_{ij}, \cdots}_\text{line process}
    )}^{\top}.
\end{equation}
These three groups of unknowns are solved with alternating optimization (see details in \textit{Supplementary Document}).
After the optimization, the new warped surface mesh $\mathcal{M}_{n}$ can be used as the initial surface to estimate the deformation field for the next frame.

\subsubsection{Topology Change Event Detection}
When detecting topology change events, we run an extra backward registration from the registered mesh to the source RGB-D image based on previous registration result, and find all cutting edges of EDG cells according to line process weights from both forward and backward registration.
There are several reasons to add this backward registration. (1) Re-using the EDG instead of resampling a new EDG from the registered mesh will preserve the correct graph node connectivity (edges along the separating boundaries having longer length due to stretching) when there is an open-to-close topology change event while the resampled EDG would not have that correct one. 
(2) It will help reducing the number of ``false positive'' cases when only considering the forward registration. ``False positive'' cases are usually caused by finding bad correspondences with outliers. This can be solved by using bidirectional correspondence search and adding backward registration follows the same way. 
(3) This backward registration is still computationally light-weight without the need to re-generate a new EDG and all computed line process weights can be directly used to guide the topology change event detection.

The formula to compute $l_{ij}$ is:
\begin{equation}
    l_{ij} = (\frac{\mu}{\mu+\|\boldsymbol{R}_{i}(\boldsymbol{g}_{i}-\boldsymbol{g}_{j})-[\boldsymbol{g}_{i}+\boldsymbol{t}_{i}-(\boldsymbol{g}_{j}+\boldsymbol{t}_{j})]\|^{2}})^{2}.
\label{eq:l}
\end{equation}
We want to set the threshold of $l_{ij}$ to distinguish between highly stretched (or compressed) edges and normal edges.
In our assumption, if the ratio of an edge stretched (or compressed) to the normal length is $20\%$, there exists a potential topology change event. 
Then a good approximation of $\mu$ is $20\% \times cell~length$.
In practice, if $l_{ij} < 0.5$ in the forward registration step and $l_{ij} < 0.8$ in the backward registration, it will be classified as a cutting edge, and there is a new topology change event detected. 

\begin{figure*}[thb]
	\centering
	\includegraphics[width=1.0\textwidth]{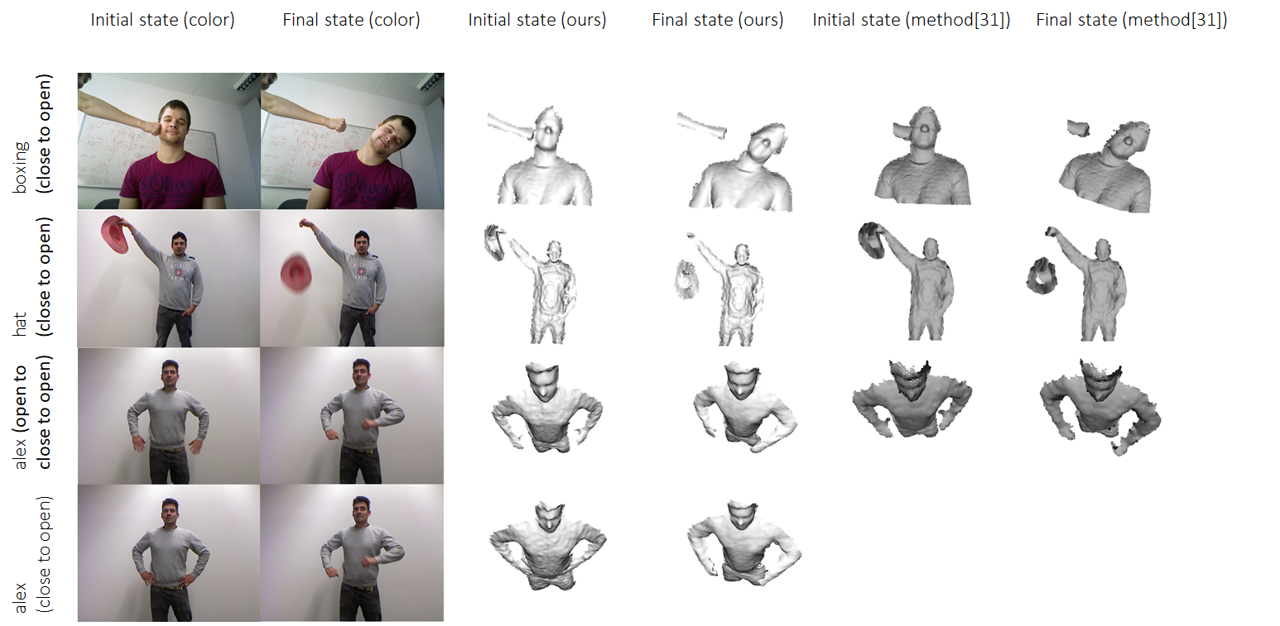}
	\caption{Effectiveness of our topology change detection on real data. Row 1 to 3 are cases shown in Zampogiannis et al's paper \cite{zampogiannis2019}. Row 4 is another challenging case from KillingFusion \cite{slavcheva2017killingfusion}.}
	\label{fig:topology_change_event}
\end{figure*}

In order to demonstrate that our topology change detection really works well, we run it on some public datasets used in~\cite{zampogiannis2019}, as shown in Fig. \ref{fig:topology_change_event}.
Our approach can also successfully detect all topology change events and update the connectivity of EDG and TSDF grids to reflect such topology changes accordingly in reconstructed geometry. It is worth noting that our method can handle a more complex case like seq ``alex (close to open)'' (from \cite{slavcheva2017killingfusion}) -- hand moving from contacting with body to no contact, which is not demonstrated in \cite{zampogiannis2019}. Besides that, Zampogiannis et al~\cite{zampogiannis2019} did not address how to reconstruct the geometry of dynamic scenes under such topology changes, as will be introduced below.

\subsubsection{Updating the Connectivity of EDG}
\begin{figure*}[htb]
	\centering
	\includegraphics[width=1.0\textwidth]{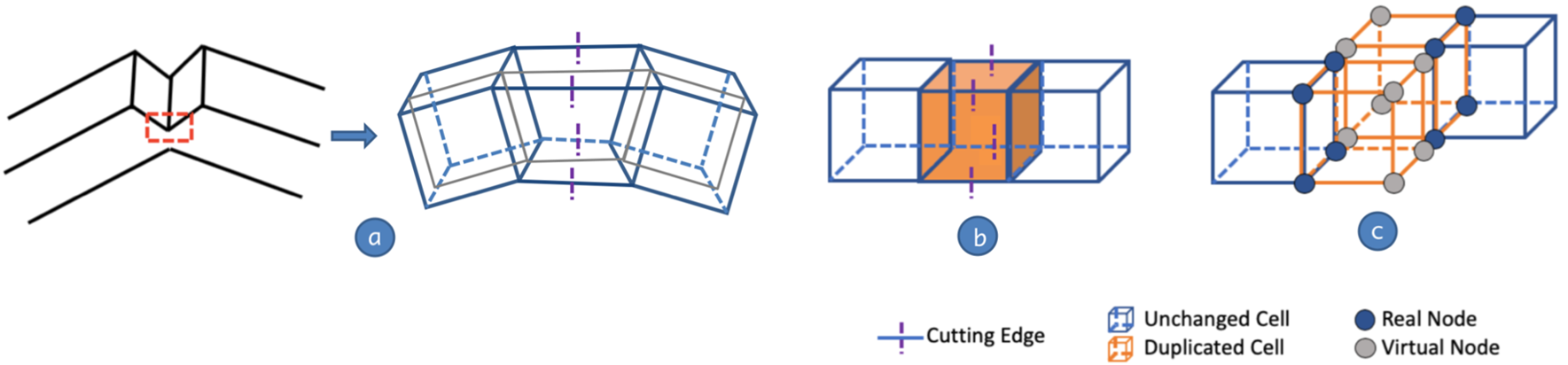}
	\caption{(a) Cutting edges (in the live frame). (b) ``To-be-duplicated'' cells found based on edge cutting (in the canonical space). (c) Final non-manifold cells (\textcolor{myorange}{orange} cells are illustrated with a small displacement to distinguish two duplicated cell which are actually at the same location).}
	\label{fig:cell_duplicate}
\end{figure*}

The most fundamental innovation in this work is to allow the cells of volumetric structure to duplicate themselves, and to allow nodes (or grid points) to have non-manifold connectivity.
In EDG $\mathcal{G}$, each cell $c^{\mathcal{G}}$ has exactly 8 nodes $\{g^{\mathcal{G}}\}$ located at its corners. 
Each node $g^{\mathcal{G}}$ can be affiliated with up to 8 cells $\{c^{\mathcal{G}}\}$ in the manifold case.
At the beginning of the 4D reconstruction, we assume all connectivity between nodes are manifold, i.e., all nodes are affiliated with 8 cells except for those on the boundary of volume.
Fig.~\ref{fig:cell_duplicate} illustrates the algorithm of our non-manifold EDG connectivity update.

\textbf{Input:} (a) A set of cutting edges detected by the method mentioned above; and (b) a set of candidate cells to be duplicated based on cutting edge detection.

\noindent\textbf{Step 1 [Cell separation]:} We separate each candidate cell $c^{\mathcal{G}}$ by removing all cutting edges and computing its connected components (CCs). 

\noindent\textbf{Step 2 [Cell duplication based on CCs]:} The candidate cells are duplicated depending on its number of CCs. 
In each duplicated cell $c^{(d)}$ we categorize its nodes into two types: (1) Real Nodes $\{g^{(r)}\}$ being those from the original cell before duplication, and (2) Virtual Nodes $\{g^{(v)}\}$ being those added to make up the duplicated cells. 
For each virtual node $g^{(v)}$, it will only be affiliated with its duplicated cell.
The transformation of each duplicated node in EDG also needs to be determined. 
For real nodes, they could inherit all properties from the original nodes.
For virtual nodes, their displacement could be extrapolated from real nodes belonging to the same cell. 
In the example of Fig.~\ref{fig:cell_duplicate}, there are 4 cutting edges on the orange cell $c^{\mathcal{G}}$ causing its 8 nodes to be separated into 2 CCs, thus the original cell $c^{\mathcal{G}}$ is replaced with 2 duplicated cells $\{c^{(d)}\}$ residing at the same location of canonical space. 

\noindent\textbf{Step 3 [Restoring connectivity]:} For any pair of geometrically adjacent duplicated cells $c^{\mathcal{G}}$ (in the canonical space), given two nodes from them respectively, merge these two nodes if:
(1) they are both real nodes and copied from the same original node, or
(2) they are both virtual nodes, copied from the same original node and connected with the same real nodes.
In the example of Fig.~\ref{fig:cell_duplicate} (c) all four nodes on the left face of the front orange cell are merged with four nodes of the left cell by the node-merging rules.

The result is shown in Fig.~\ref{fig:cell_duplicate} (c). After restoring the connectivity, the final EDG has been fully assembled, respecting the topology change of the target RGB-D image.
After a few edge cutting and cell duplication operations, the connectivity of nodes will become non-manifolds.

\subsection{Topology-Change-Aware Geometric Fusion}
Now we describe how to update and fuse the TSDF volume based on the deformation field estimated from the previous step and the depth image $\mathcal{D}_{n}$ in the $n^{th}$ frame.
In order to accelerate the registration running speed and improve the reconstruction quality of geometry, a strategy of multi-level grids is employed in this paper.
The resolution of EDG is typically lower than that of TSDF volume, with a ratio of $1:(2k+1)$ in each dimension ($k \in \{1,2,3\}$ in our experiments).
Thus, care needs to be taken when updating the connectivity of TSDF volume if the resolution of TSDF volume grid is different from that of EDG.

\subsubsection{Updating TSDF Volume}
\begin{figure}[htb]
	\centering
	\includegraphics[width=0.54\textwidth]{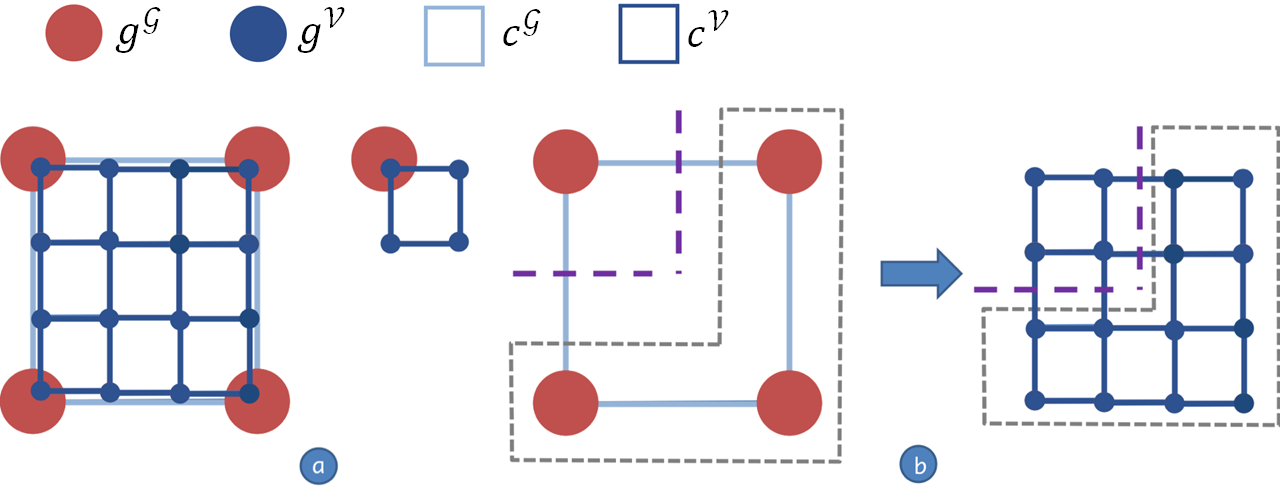}
	\includegraphics[width=0.42\textwidth]{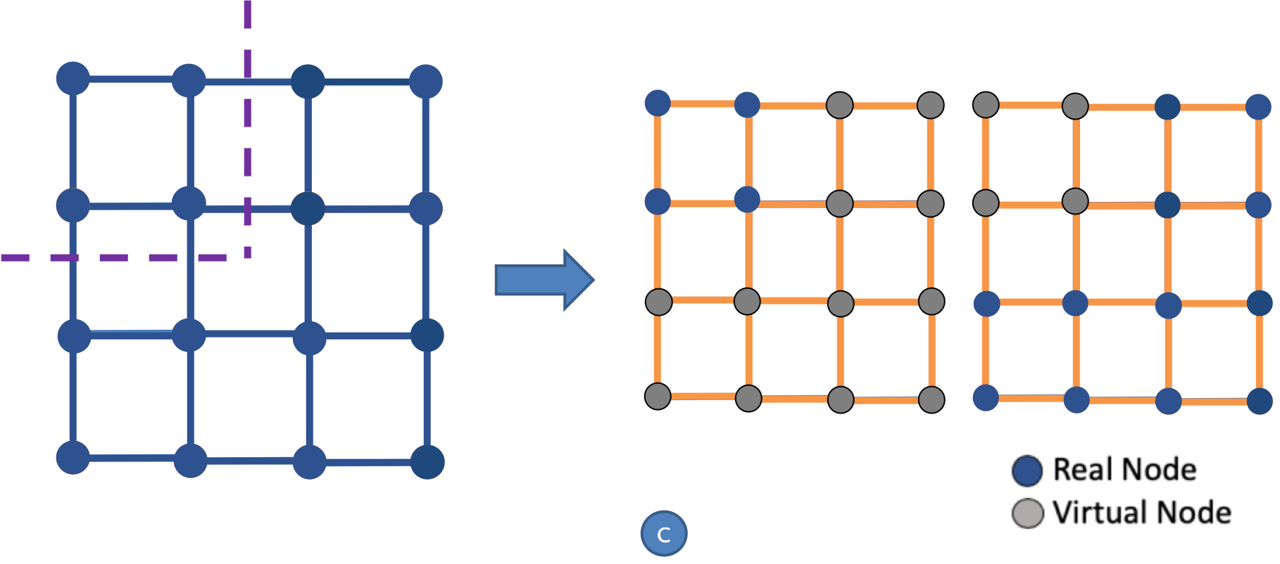}
	\caption{(a) A cell $c^{\mathcal{G}}$ of EDG, its embedded TSDF volume cells $\{c^{\mathcal{V}}\}$ and a set of voxels $\{g^{\mathcal{V}}\}$ belonging to a node $g^{\mathcal{G}}$ of this cell. (b) Connectivity  propagation from an EDG cell to its embedded TSDF volume cells. (c) Connectivity update of TSDF volume.}
	\label{fig:tsdf_cell_duplicate}
\end{figure}

Once the deformation field is estimated, the connectivity of EDG should be propagated to TSDF volume and the depth image should be fused as well.
Fig.~\ref{fig:tsdf_cell_duplicate} shows key steps on how to propagate the connectivity to TSDF volume.

\textbf{Input:} (a) EDG cells and their embedded TSDF volume cells; and (b) a set of cutting edges in EDG.

\noindent\textbf{Step 1 [Cell separation]:} Each EDG cell contains $(2k+1)^{3}$ TSDF cells and $(2k+2)^{3}$ TSDF voxels. Each EDG node controls $(k+1)^{3}$ voxels.
Fig.~\ref{fig:tsdf_cell_duplicate} (a) shows a 2D case when $k = 1$.
We separate each volume cell $c^{\mathcal{V}}$ by considering the connected components (CCs) of its associated EDG cell -- the CCs belonging of each voxel is the same as its associated EDG node.
If two vertices of an edge belong to different CCs, this edge is treated as a cutting edge(Fig.~\ref{fig:tsdf_cell_duplicate} (b)).

\noindent\textbf{Step 2 [Cell duplication based on CCs]:} TSDF volume cells are duplicated depending on the number of CCs of an EDG cell $c^{\mathcal{G}}$, as shown in Fig.~\ref{fig:tsdf_cell_duplicate} (c). 
Therefore, even though the number of CCs of TSDF volume cell on the top left is $1$, it will still be duplicated as two copies: one copy containing all real nodes while the other copy containing all virtual nodes. 

\begin{figure*}[htb]
	\centering
	\includegraphics[width=1.0\textwidth]{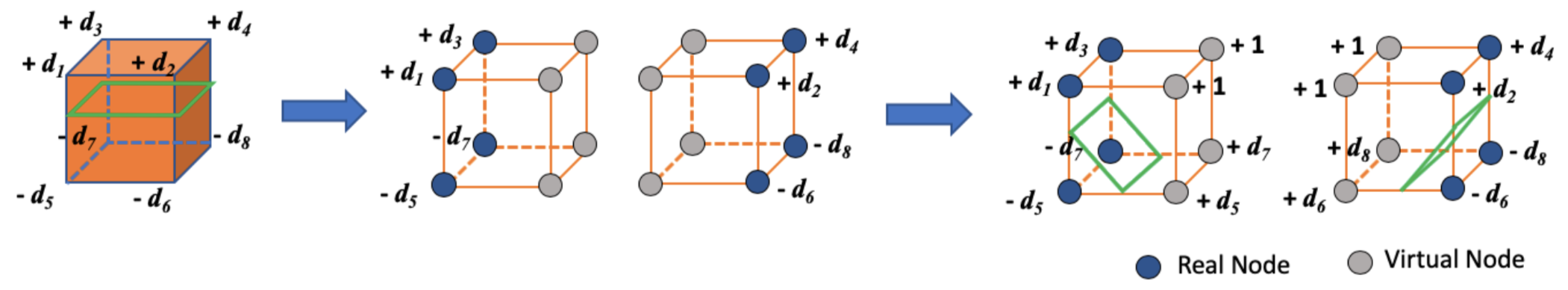}
	\caption{The updating rule for the signed distances on virtual nodes of TSDF grids. The green surfaces denote the zero crossing surface inside the cells.}
	\label{fig:tsdf_update}
\end{figure*}

For those virtual nodes in the TSDF volumetric structure, their TSDF values need to be updated with caution. 
Here we use the following three updating rules: (1) For all real nodes, since we need to keep the continuity of their TSDF, we directly inherit their TSDF value from the original cell. (2) For all virtual nodes that are connected to real nodes, if their connected real node has negative TSDF value (meaning inside the surface), we set the TSDF of the corresponding virtual node by negating that value, i.e. $-d \rightarrow +d$. (3) For all remaining virtual nodes that have not been assigned TSDF values, we simply set their values as $+1$. Fig.~\ref{fig:tsdf_update} shows an illustration of these TSDF updating rules. Note that all these TSDF values might continue to be updated by the depth fusion step that follows. 

\noindent\textbf{Step 3 [Restoring connectivity]:} For any pair of geometrically adjacent duplicate cells $c^{\mathcal{V}}$ (in the canonical space), given two nodes $g^{\mathcal{V}}$ from them respectively, the merging rule is a bit different from the one used for EDG cell $c^{\mathcal{G}}$.
We merge two nodes $g^{\mathcal{V}}$ if they are copied from the same original node and they are: (1) both real nodes, or (2) both virtual nodes.

Because the connectivity update of EDG is propagated to the TSDF grid, the geometry represented by TSDF could reflect topology changes and each cell $c^{\mathcal{V}}$ in the volume could find its correct EDG cell association.
Next, all voxels will be warped to the live frame by the estimated deformation field. Similar to~\cite{newcombe2015dynamicfusion},  depth information is fused into the volume in the canonical space.

\subsubsection{Preparing for the Next Frame}
In order to guide the estimation of deformation field for the next coming frame, we need to extract a surface mesh from the TSDF volume in the canonical space.
Since the TSDF volumetric grid could become non-manifold, the marching cubes method needs to be modified to make it adapted to the topology changes.

\textbf{Extended marching cubes method:}
In the classic fusion framework, each TSDF volume cell is unique. Given the position of the left-front-bottom voxel in the canonical frame, the only corresponding EDG/TSDF grid cell is returned in $O(1)$ time. Now because of cell duplication, this rule will not hold.
Therefore, for each voxel, we also store cell information.
For each EDG node, we just need to store the id of its belonged EDG cell.
For TSDF volume, we do not want to maintain another list of all volume cells.
We directly store the list of voxel ids for one specific volume cell -- the cell having this voxel as its left-front-bottom voxel.
There are two benefits brought by adding this extra information: (1) it will help identifying the corresponding TSDF volume cell for every voxel once cells are duplicated;
(2) after extracting the surface mesh by marching cubes method, each vertex also inherits the id of its belonged EDG cell, which makes it convenient to warp the mesh according to the deformation field defined by EDG.
Finally, we extract triangle mesh for each TSDF volumetric cell in parallel and merge vertices on shared edges between cells.

\textbf{Expanding EDG:}
As the 3D model grows by fusion of new geometry, the support of deformation field -- EDG should also be expanded.
Because we have a predefined grid structure for EDG and the primitive element of our EDG connectivity update algorithm is EDG cell, 
different from other fusion-based methods, we directly activate those EDG cells which embed the newly added geometry part to maintain the ability to separate and duplicate cells when there are new topology changes.

\section{Experimental Results}
There are several specific public datasets on topology change problems. 
Tsoli and Argyros~\cite{tsoli2016tracking} provided both synthetic and real data, from which the synthetic data is generated through physics-based simulation in Blender and the real data is captured with Kinect v2.
Slavcheva \textit{et al.}~\cite{slavcheva2017killingfusion} also published their data.
We evaluate qualitatively and quantitatively our method based on those mentioned datasets and the experimental results from the authors.
Then ablation study is included to show the effect of different key components in our entire pipeline.

\subsection{Evaluation on Synthetic Data}
The baseline methods we select for synthetic data evaluation are CPD~\cite{myronenko2010point}, MFSF~\cite{garg2013variational}, Tsoli and Argyros's method~\cite{tsoli2016tracking} and VolumeDeform~\cite{innmann2016volumedeform}.
The first three methods are template based non-rigid registration methods.
Specifically, Tsoli and Argyros's method can deal with deformable surfaces that undergo topology changes.
VolumeDeform and our method are both template-less fusion-based reconstruction methods.
DynamicFusion~\cite{newcombe2015dynamicfusion} has bad performance on this synthetic dataset because it cannot deal well with deformations parallel to camera screen, so we do not compare with it.

\begin{figure*}[htb]
    \centering
    \includegraphics[width=1.0\textwidth]{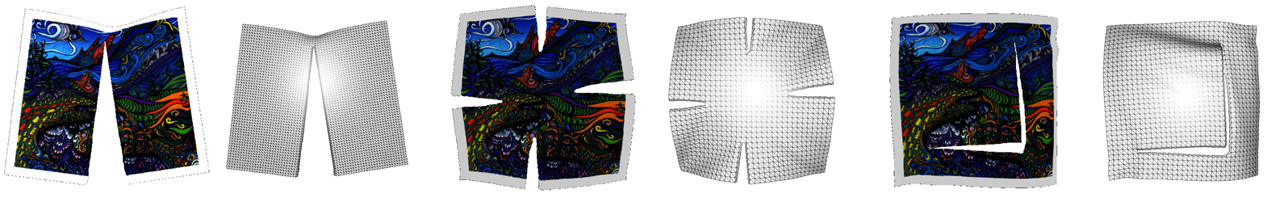}
    \caption{Our reconstruction results on Tsoli and Argyros's synthetic dataset: seq1, seq2, and seq3, from left to right.}
    \label{fig:teaser_image}
\end{figure*}

We select two metrics proposed in Tsoli and Argyros's paper~\cite{tsoli2016tracking}: (1) Euclidean distance from ground truth; and (2) the number of vertices off the surface. 
We believe metric 1 can quantitatively evaluate the overall reconstruction quality while metric 2  provides a deeper insight about how the topologically changed parts are reconstructed.
There will be lots of vertices ``off the surface'' if the topologically changed part is not well considered and processed.
We refer the readers to Tsoli and Argyros's paper~\cite{tsoli2016tracking} for detailed definition of these metrics.
Here, the distance measurement for both metrics are expressed as a percentage of the cell width of the underlying grid.
Because VolumeDeform and our method are reconstruction methods without any pre-defined template, to be consistent with Tsoli and Argyros's experiment, we allocate the volume according to the same grid cell width and the resolution of their template in x and y axis directions.
\begin{figure}[htb]
    \centering
    \subfloat[]{\includegraphics[width=0.6\textwidth]{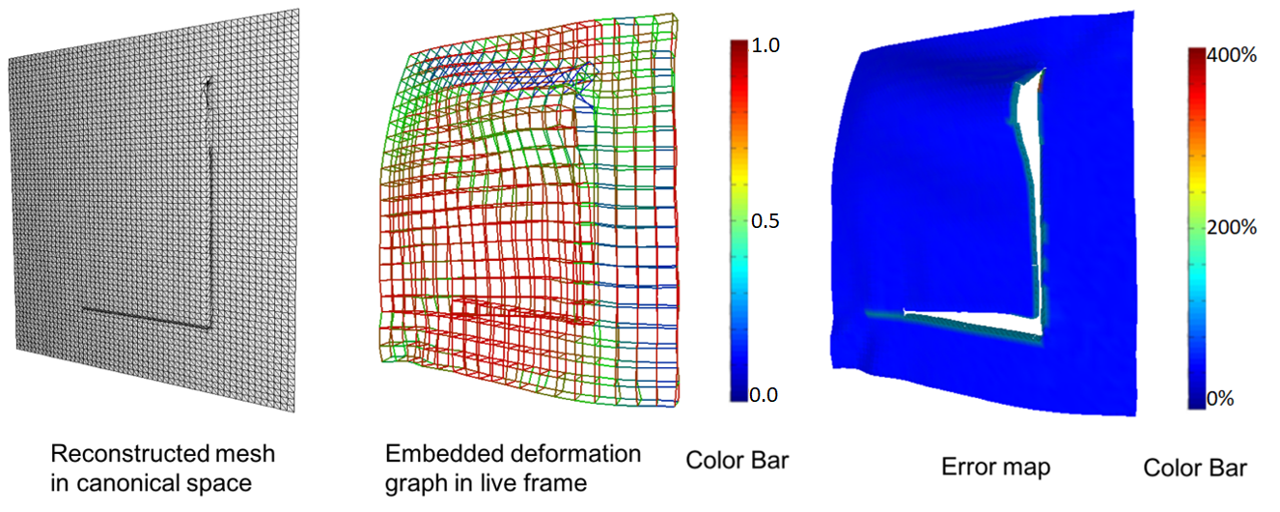}}
    \subfloat[]{\includegraphics[width=0.25\textwidth]{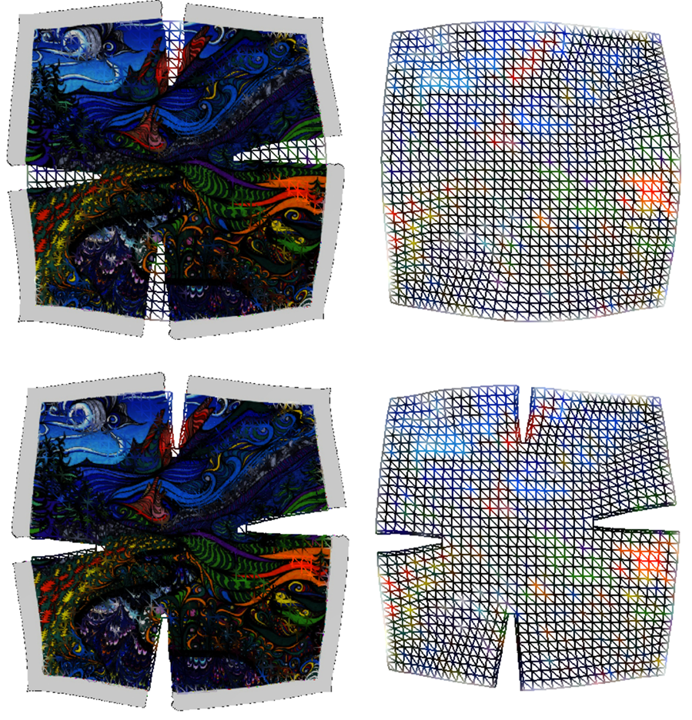}}
    \caption{(a) Qualitative comparison of our reconstructed seq3 data with the ground truth. 
    (b) Reconstruction results on seq2 of Tsoli and Argyros's dataset by VolumeDeform (top row) and our method (bottom row).}
    \label{fig:syn_data_qualitative}
\end{figure}

Fig.~\ref{fig:syn_data_qualitative} (a) shows a reconstruction result on frame \#36 of seq3 in Tsoli and Argyros's dataset.
The color \textcolor{red}{\emph{Red}}-\textcolor{green}{\emph{Green}}-\textcolor{blue}{\emph{Blue}} on the EDG edge represents line process weights $l_{ij}$ from 1 to 0. The error map using the color-bar on the right shows the Euclidean distance from ground truth, expressed as the percentage of the cell width in TSDF volume.
We can see that the reconstructed mesh in live frame reflects the topology change in this case and so does the reconstructed mesh in canonical space.
The line process weights of edges also represent the presence of deformation discontinuity. 
\begin{figure}[th]
    \centering
    \includegraphics[width=0.99\textwidth]{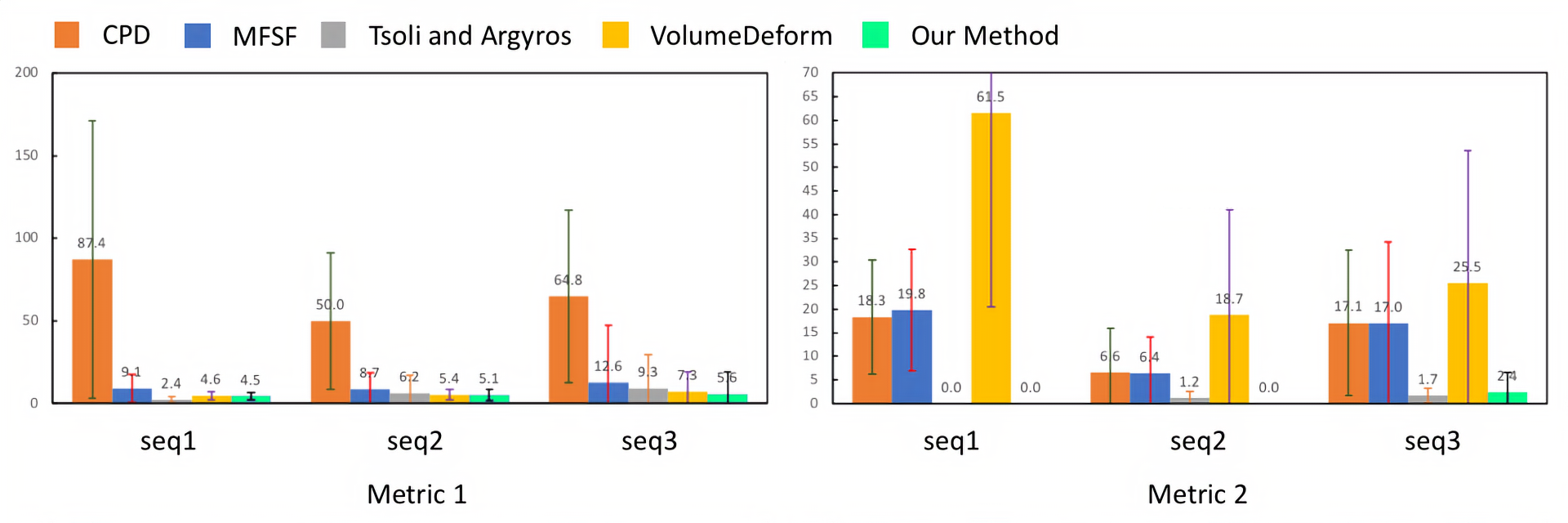}
    \caption{Quantitative comparison with other methods. Metric 1: Euclidean distance from ground truth. Metric 2: number of vertices off the surface.}
    \label{fig:syn_data_metric}
\end{figure}

We evaluate all five methods on synthetic dataset: a single cut (seq1), multiple non intersecting cuts (seq2), two intersecting cuts (seq3) (Fig.~\ref{fig:teaser_image}).
Fig.~\ref{fig:syn_data_metric} show the performance of each method based on the two error metrics. Our method outperforms all other methods on seq2 and seq3 in terms of the distance from ground truth.
Only Tsoli and Argyros's method does a better job on seq1 than ours.
Under metric 2, our method outperforms all other methods on seq2.
On seq1, our method is better than all other methods except Tsoli and Argyros's method.
On seq3, our method has a bit higher average error than Tsoli and Argyros's method.
Fig.~\ref{fig:syn_data_qualitative} (b) displays the reason why VolumeDeform performs well under metric 1 but much worse under metric 2. 
It is because VolumeDeform keeps a fixed-topology grid structure to represent the deformation field and the geometry, and has no mechanism to deal with topology changes.

\subsection{Comparison to State-of-the-art on Real Data}
\begin{figure*}[htb]
	\centering
	\includegraphics[width=1.0\textwidth]{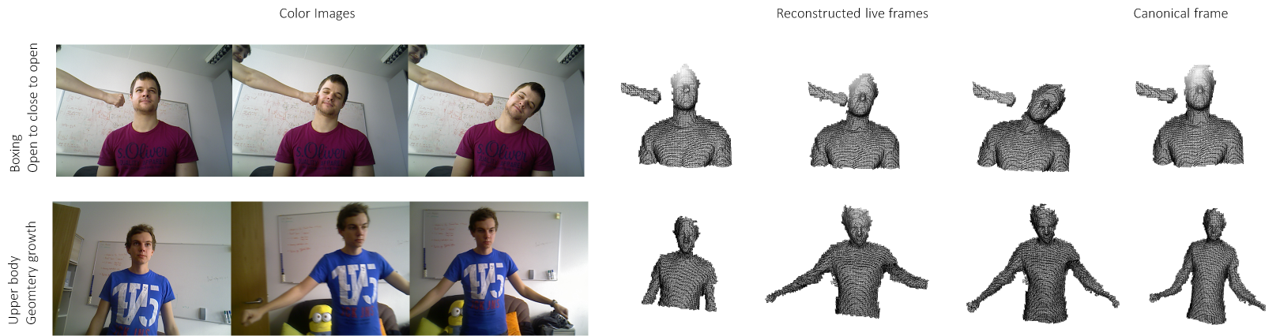}
	\caption{Reconstruction results on real open-to-close and geometry growth data. Top row: open-to-close case; bottom row: geometry growth on body and arms.
	}
	\label{fig:real_data}
\end{figure*}

Our method inherits from the classic DynamicFusion \cite{newcombe2015dynamicfusion} framework, so two characteristics of DynamicFusion are kept: (1) open-to-close motions can be solved very well and (2) geometry will grow as more regions are observed during the reconstruction.
Fig. \ref{fig:real_data} shows some reconstruction results on VolumeDeform datasets. In the boxing sequence, some key frames reconstruction results illustrate that our method works well on an open-to-close-to-open motion. In the second sequence, the reconstructed geometry of upper body is rendered from a different viewpoint to make it easier to see the geometry growth during fusion. 

\begin{figure*}[htb]
	\centering
	\subfloat[]{\includegraphics[width=0.20\textwidth]{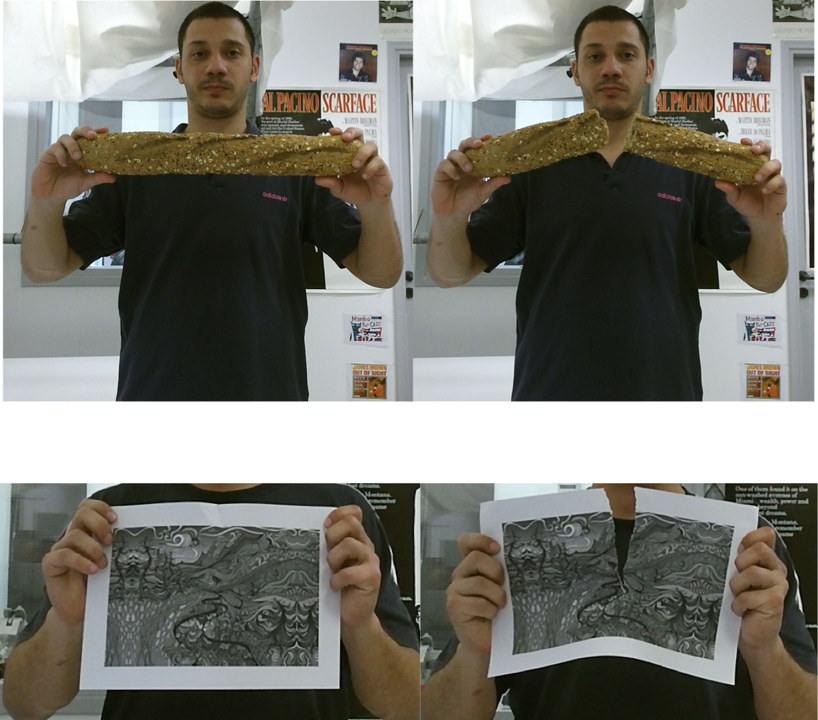}}
	\subfloat[]{\includegraphics[width=0.25\textwidth]{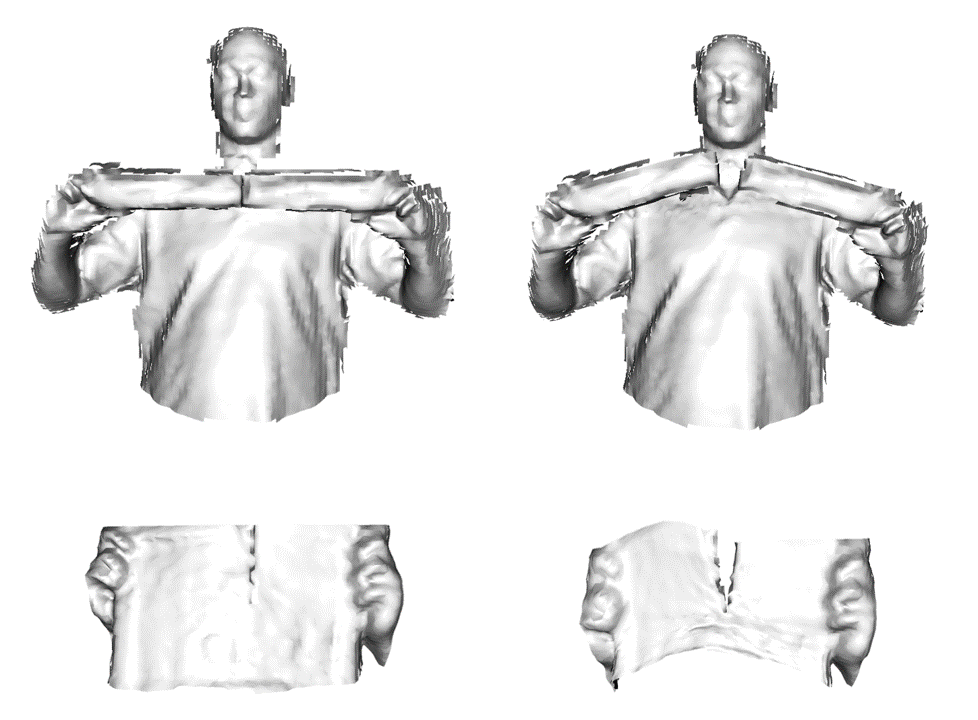}}
	\subfloat[]{\includegraphics[width=0.26\textwidth]{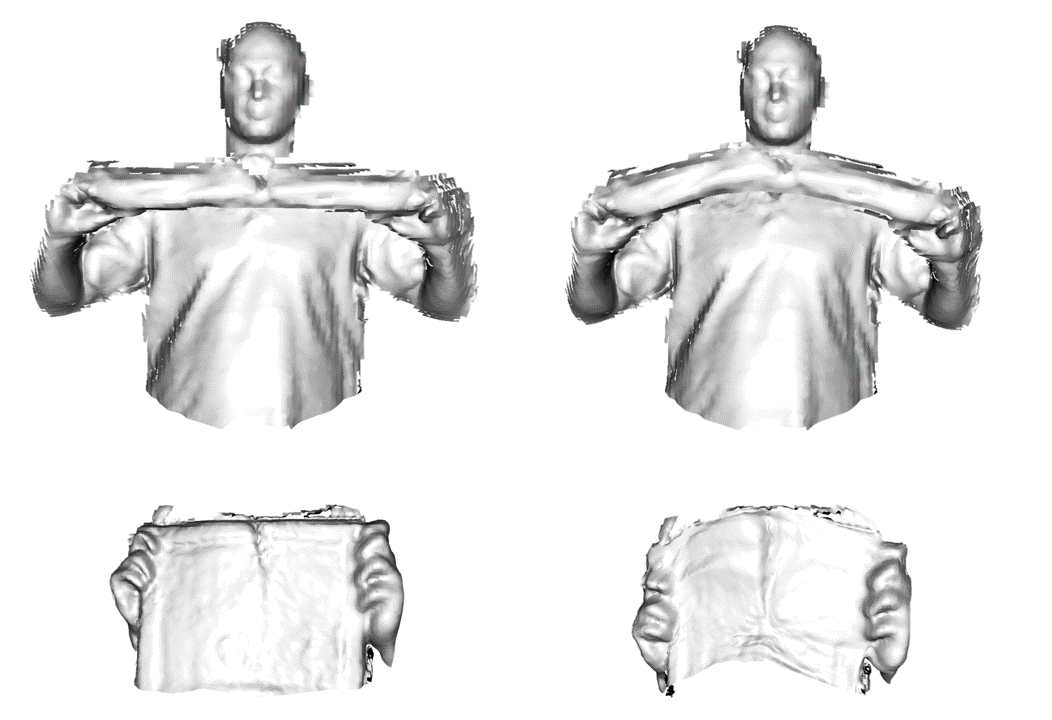}}
	\subfloat[]{\includegraphics[width=0.27\textwidth]{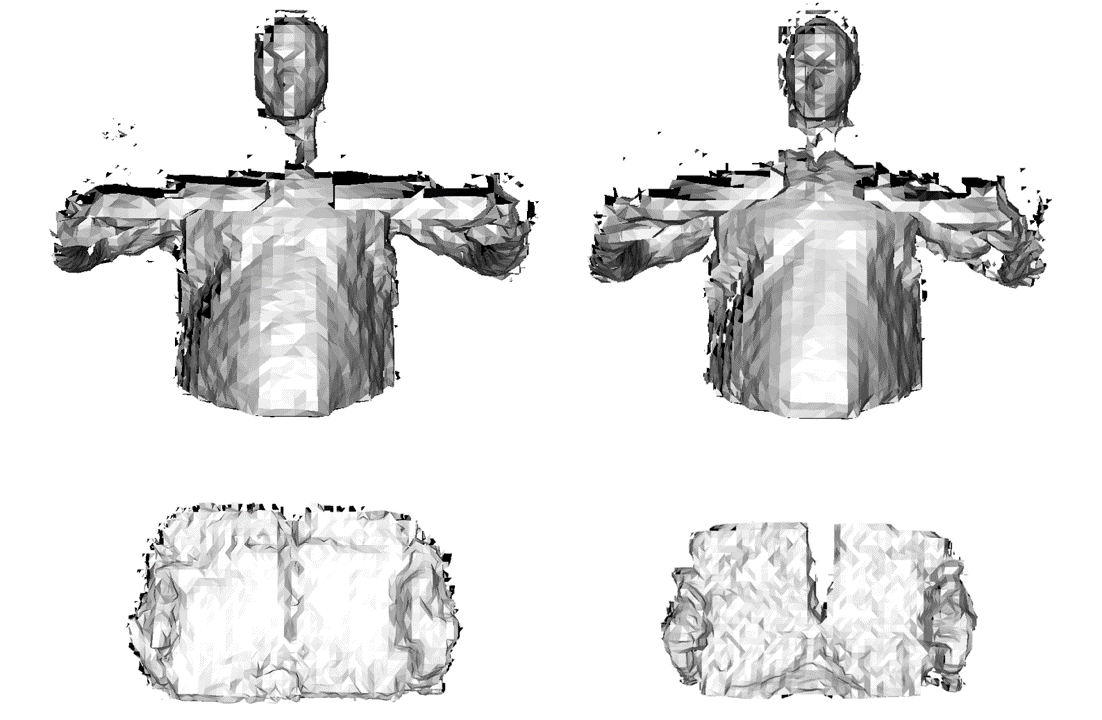}}
	\caption{Results on real data with topology changes. From left to right: (a) starting frame and live frame; (b) reconstructed geometry in canonical frame and live frame by our method; (c) VolumeDeform; (d) KillingFusion.
	}
	\label{fig:more_real_data}
\end{figure*}

The methods we compare for real data are VolumeDeform~\cite{innmann2016volumedeform} and KillingFusion~\cite{slavcheva2017killingfusion}. Fig.~\ref{fig:more_real_data} shows such comparison, where the first row is a bread breaking sequence and the second row is a paper tearing sequence.
The leftmost a couple of images are RGB images for reference: images of starting frame and current live frame.
The remaining 3 pairs of images show the reconstruction results by our method, VolumeDeform and KillingFusion.
We can see that VolumeDeform could not update geometry correctly while both KillingFusion and our method could handle topology changes.
But we can see that KillingFusion produces less smooth reconstructed surfaces compared to ours, even though all three methods use the same resolution of TSDF volume.
The entire reconstructed sequences shown in Fig. \ref{fig:topology_change_event}, \ref{fig:teaser_image}, \ref{fig:real_data}, \ref{fig:more_real_data} are in the \textit{Supplementary Video}.

\subsection{Ablation Study}
\begin{figure}[htb]
	\centering
	\subfloat[]{\includegraphics[width=0.16\textwidth]{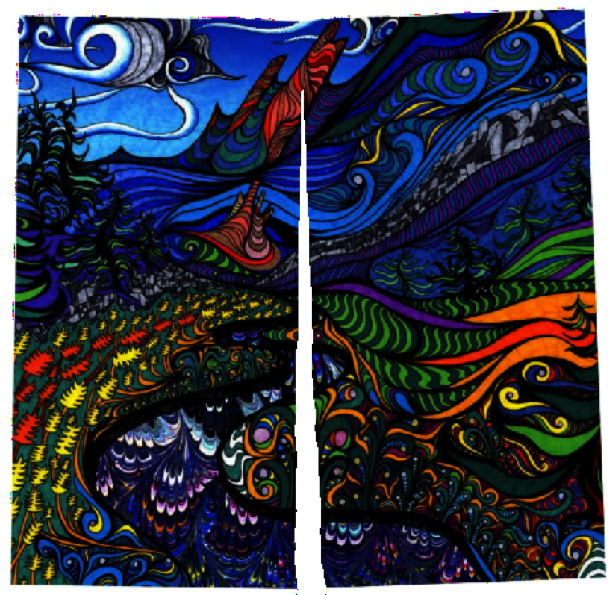}\label{fig:line_process:a}}
	~\subfloat[]{\includegraphics[width=0.17\textwidth]{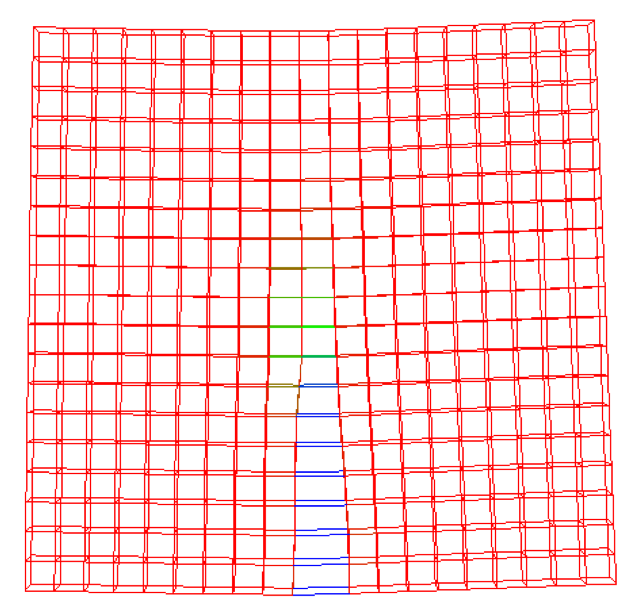}\label{fig:line_process:b}}
	~\subfloat[]{\includegraphics[width=0.165\textwidth]{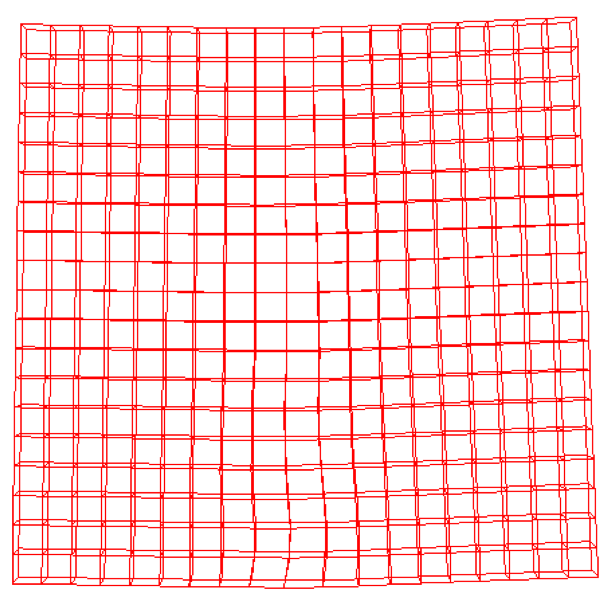}\label{fig:line_process:c}}
	\caption{Effect of line process: (a) target point cloud, (b) with line process, (c) without line process. Color \textcolor{red}{\emph{Red}}-\textcolor{green}{\emph{Green}}-\textcolor{blue}{\emph{Blue}} on the edge means $l_{ij}$ from 1 to 0.}
	\label{fig:line_process}
\end{figure}

\textbf{Effect of line process based registration:}
Fig.~\ref{fig:line_process} shows the comparison of registration results with/without line process in the ARAP regularity term. 
It could be noted that Fig.~\ref{fig:line_process} (b) has better registration result than Fig.~\ref{fig:line_process} (c) in the tearing part. 
The line process weights in Fig.~\ref{fig:line_process} (b) also indicate the discontinuity of edges which help identifying cutting edges given a threshold.
\vspace{-3.0mm}

\begin{figure*}[htb]
	\centering
	\includegraphics[width=0.48\textwidth]{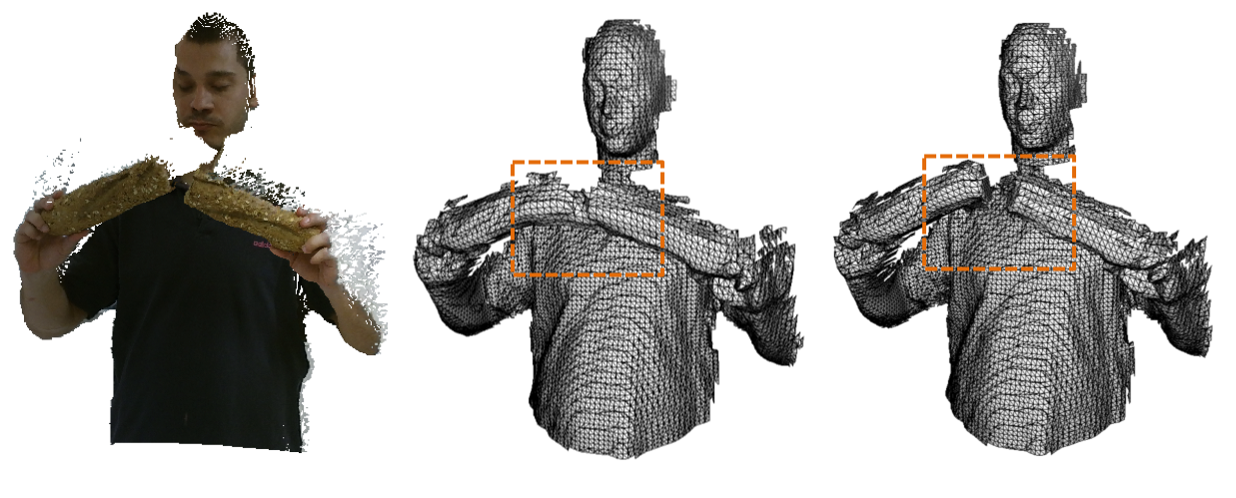}
	\includegraphics[width=0.48\textwidth]{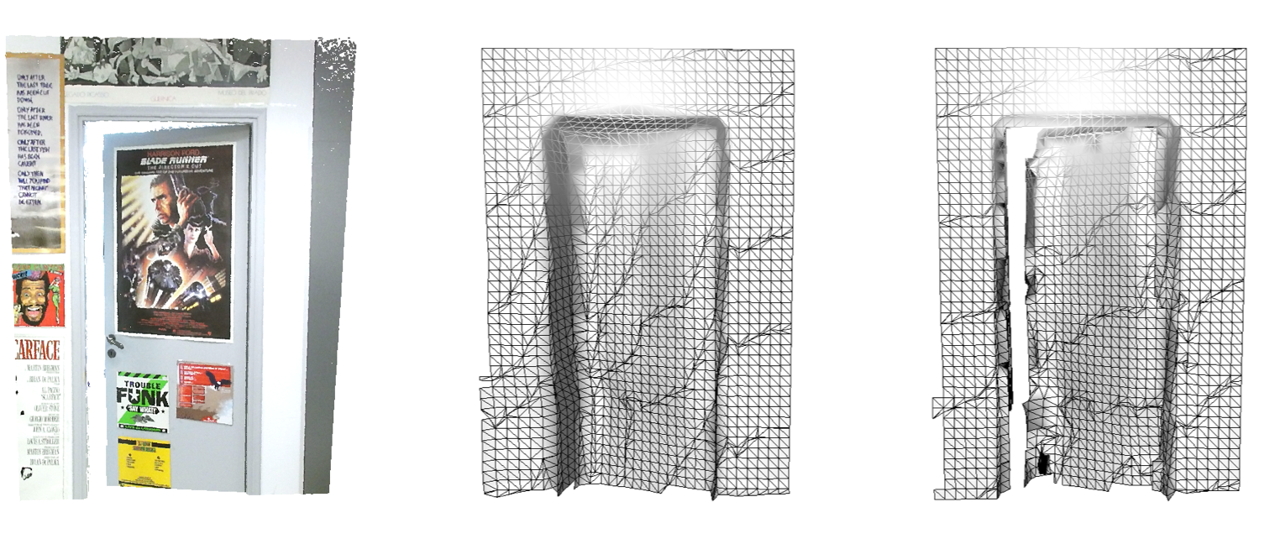}
	\caption{Effect of connectivity update. Left: input point cloud. Middle: result without connectivity update. Right: result with connectivity update.}
	\label{fig:connect_update}
\end{figure*}

\textbf{Effect of connectivity update:}
Fig.~\ref{fig:connect_update} demonstrates the effect of connectivity update. 
Without the connectivity update, topology changes will not be correctly reconstructed even though our topology-change-aware registration could help aligning surface towards the target point cloud.

\textbf{Effect of different resolutions:}
As previous work points out (Fig. 10 in \cite{innmann2016volumedeform}), higher resolution of TSDF volume results in better reconstructed details and vice versa. This is a common issue of all fusion-based reconstruction, and so is our algorithm. Due to the assumption of all cutting edges being cut in mid-points, lower resolution of EDG may cause inaccurate cutting positions. However, we have two ways to alleviate such an effect: 1) Increasing the resolution of EDG; 2) Our multi-level grids and connectivity propagation algorithm. Moreover, although EDG may have a lower resolution but a higher resolution of TSDF can complement this by reconstructing more detailed geometry. In the bread breaking and paper tearing sequences, the voxel resolution is 6mm while cell resolution is 30mm.

\section{Conclusion and Future Work}
In this paper we introduce a new topology-change-aware fusion framework for 4D dynamic scene reconstruction, by proposing the non-manifold volumetric grids for both EDG and TSDF, as well as developing an efficient approach to estimate a topology-change-aware deformation field and detect topology change events.
Our method also has some limitations.
One failure case caused by mid-point cutting assumption is cloth tearing with complex boundary. A lower resolution EDG tends to make the tearing boundary towards a line. There also exists other topology cases that our method is not designed to handle such as surface merging cases from genus 0 to higher genus, e.g. a ball morphs to a donut.
Our system currently runs at around $5$ FPS. But our system design is oriented towards parallel computation, as discussed in the \textit{Supplementary Document}.
In the future, we would like to perform code optimization and fully implement it in CUDA to achieve real-time performance.

~\\
\noindent \textbf{Acknowledgement} 
This research is partially supported by National Science Foundation (2007661). The opinions expressed are solely those of the authors, and do not necessarily represent those of the National Science Foundation.
\clearpage
%
%
\bibliographystyle{splncs04}
\bibliography{egbib}
\end{document}


\pagestyle{headings}
\mainmatter
\def\ECCVSubNumber{2540}  

\title{Topology-Change-Aware Volumetric Fusion for Dynamic Scene Reconstruction\\
\textemdash Supplementary Material\textemdash} 

\titlerunning{Supplementary Material}
%
\author{Chao Li \and Xiaohu Guo}
\authorrunning{C. Li and X. Guo}
%
\institute{Department of Computer Science,\\
	The University of Texas at Dallas\\
	\email{ \{Chao.Li2, xguo\}@utdallas.edu}
	}
\maketitle

\begin{table*}[t]
	\centering
	\caption{Definitions of the math symbols used in this paper.}
	\label{table:symbols}
\begin{adjustbox}{width=\columnwidth,center}
\begin{tabular}{|l|l|}
\hline
$\mathcal{C}_{n}$:color image of the $n^{th}$ frame 
& $\mathcal{V}_{n}$:TSDF grid updated from the $n^{th}$ frame\\ \hline

$\mathcal{D}_{n}$:depth image of the $n^{th}$ frame
& $\mathcal{G}_{n}$:EDG grid updated from the $n^{th}$ frame\\ \hline

$e_{ij}$:edge of EDG connecting the $i^{th}$ and $j^{th}$ nodes
& $l_{ij}$:line process parameter between the $i^{th}$ and $j^{th}$ nodes \\
\hline

$\{c^{\mathcal{V}}\}$:the cells of TSDF volume
& $\{c^{\mathcal{G}}\}$:the cells of EDG\\ \hline

$\{g^{\mathcal{V}}\}$:the grid-points (voxels) of TSDF volume
& $\{g^{\mathcal{G}}\}$:the nodes of EDG\\ \hline

\multicolumn{2}{|l|}{$\{\mathbf{R},\mathbf{t}\}$:global rotation and translation from the canonical to current frame}\\ \hline

\multicolumn{2}{|l|}{$\{\mathbf{R}_{i},\mathbf{t}_{i}\}$:local rotation and displacement of the $i^{th}$ node from the canonical to current frame} \\ \hline

\multicolumn{2}{|l|}{$\overline{\mathcal{M}_{n}}$:surface mesh defined in the canonical space and reconstructed from the images of first $n$ frames}  \\ \hline

\multicolumn{2}{|l|}{$\mathcal{M}_{n}$:surface mesh of $\overline{\mathcal{M}_{n}}$ being warped to the space of the $n^{th}$ frame}  \\ \hline

\end{tabular}
\end{adjustbox}
\end{table*}
This supplementary file consists of:
\begin{itemize}
  \item Details of optimization method
  \item Details of data structure design towards real-time performance
\end{itemize}

\section{Details of Optimization Method}
For the ease of discussion, we provide all notations of the math symbols of this paper in Table~\ref{table:symbols}. 

Alternating optimization: solve three groups of unknowns by fixing the other two groups and solve one group.

\noindent \textbf{Step 1 [Fix $\{\boldsymbol{R}^{\top}_{i}\}$ and $\{l_{ij}\}$, solve $\{\boldsymbol{t_{i}}\}$] }
Set
\begin{equation}
    \frac{\partial E_{total}(\boldsymbol{t_{i}})}{\partial \boldsymbol{t_{i}}} = \boldsymbol{0}
\label{eq:deriv}
\end{equation}
Solve $A^{\top}WA\boldsymbol{x} = -A^{\top}W\boldsymbol{b}$ with Preconditioned Conjugate Gradient (PCG), where $\boldsymbol{x}$ is the stacked vector of all $\boldsymbol{t_{i}}$ and W is the diagonal matrix of term weights.
\begin{equation}
A=\begin{pmatrix}
 & & \cdots& & \\
 \cdots& \alpha_{i}(\boldsymbol{R}^{\top}\boldsymbol{n}_{y})^{\top}& \cdots& \alpha_{j}(\boldsymbol{R}^{\top}\boldsymbol{n}_{y})^{\top}& \cdots& \\
 & & \cdots& & \\
 \cdots& \alpha_{i}\boldsymbol{R}& \cdots& \alpha_{j}\boldsymbol{R}& \cdots& \\
 & & \cdots& & \\
 \cdots& l_{ij}I& \cdots& -l_{ij}I& \cdots& \\
 & & \cdots& & \\
\end{pmatrix}
\label{eq:A}
\end{equation}

\begin{equation}
b = \begin{pmatrix}
 \vdots \\
 \boldsymbol{n}^{\top}_{y}(\boldsymbol{T}(\boldsymbol{x})-\boldsymbol{y}) \\
 \vdots\\
 \boldsymbol{T}(\boldsymbol{f})-\boldsymbol{y} \\
 \vdots \\
 l_{ij}(\boldsymbol{R}_{i}-I)(\boldsymbol{g}_{i}-\boldsymbol{g}_{j}) \\
 \vdots \\
\end{pmatrix}
\label{eq:b}
\end{equation}

\noindent \textbf{Step 2 [Fix $\{\boldsymbol{t_{i}}\}$ and $\{l_{ij}\}$, solve $\{\boldsymbol{R}^{\top}_{i}\}$] }
For each $\boldsymbol{R}^{\top}_{i}$, it is a least square rigid estimation, which has closed form solution. 
Therefore, all $\{\boldsymbol{R}^{\top}_{i}\}$ could be solved in parallel.

First, compute the cross-covariance matrix $A$ for all $\boldsymbol{g}_{i}$ corresponding terms:
\begin{equation}
A = XLY^{\top}
\label{eq:coA}
\end{equation}

\begin{equation}
X = \begin{pmatrix}
\cdots \\
\boldsymbol{g}_{i} - \boldsymbol{g}_{j} \\
\cdots \\
\end{pmatrix}
\label{eq:X}
\end{equation}

\begin{equation}
L = \begin{pmatrix}
\ddots& & \\
& l_{ij}& \\
& & \ddots \\
\end{pmatrix}
\label{eq:L}
\end{equation}

\begin{equation}
Y = \begin{pmatrix}
\cdots \\
[\boldsymbol{g}_{i}+\boldsymbol{t}_{i} - ( \boldsymbol{g}_{j}+\boldsymbol{t}_{j})]^{\top} \\
\cdots \\
\end{pmatrix}
\label{eq:Y}
\end{equation}

Secondly, by solving the Singular Value Decomposition (SVD) of matrix $A$, the optimal value of $\Delta R^{*}_{i}$ is:
\begin{equation}
\Delta R^{*}_{i}=V\begin{pmatrix}
1&  &  \\ 
&  1&  \\ 
&  &  det(VU^{\top})\\ 
\end{pmatrix}U^{\top},
\label{eq:R}
\end{equation}
where
\begin{equation}
{A} = U \Sigma V^{\top},
\label{eq:S}
\end{equation}

\noindent \textbf{Step 3 [Fix $\{\boldsymbol{R}^{\top}_{i}\}$ and $\{\boldsymbol{t_{i}}\}$, solve $\{l_{ij}\}$] }
By setting 
\begin{equation}
    \frac{\partial E_{reg}(l_{ij})}{\partial l_{ij}} = \boldsymbol{0}
\label{eq:l_deriv}
\end{equation},
we have
\begin{equation}
    l_{ij} = (\frac{\mu}{\mu+\|\boldsymbol{R}_{i}(\boldsymbol{g}_{i}-\boldsymbol{g}_{j})-[\boldsymbol{g}_{i}+\boldsymbol{t}_{i}-(\boldsymbol{g}_{j}+\boldsymbol{t}_{j})]\|^{2}})^{2}
\label{eq:l}
\end{equation}.

\noindent \textbf{Initialization:} $\boldsymbol{R}^{\top}_{i} \leftarrow \boldsymbol{I}, \boldsymbol{t_{i}} \leftarrow \boldsymbol{t_{i}'}$(optimal $\boldsymbol{t_{i}}$ solved from previous frame), $l_{ij} \leftarrow 1.0$

\section{Details of Data Structure Design Towards Real-Time Performance}
There are several requirements to meet when re-designing the data structure of our topological-change-aware fusion framework towards real-time performance.
\begin{enumerate}
    \item Efficient cell duplicate and merge operation.
    \item Fast EDG/volume cell to node/voxel mapping and reverse mapping when duplicate cell exists.
\end{enumerate}
For the second requirement, in details, fast vertex/voxel to EDG cell mapping and EDG cell to node mapping is required to compute the deformation of each vertex/voxel by trilinear interpolation based on the estimated deformation field.
Fast volume cell to voxel mapping is required to do marching cubes to extract surface mesh.

\subsection{Embedded Deformation Graph (EDG)}
\begin{figure}[th]
	\centering
	\includegraphics[width=0.5\textwidth]{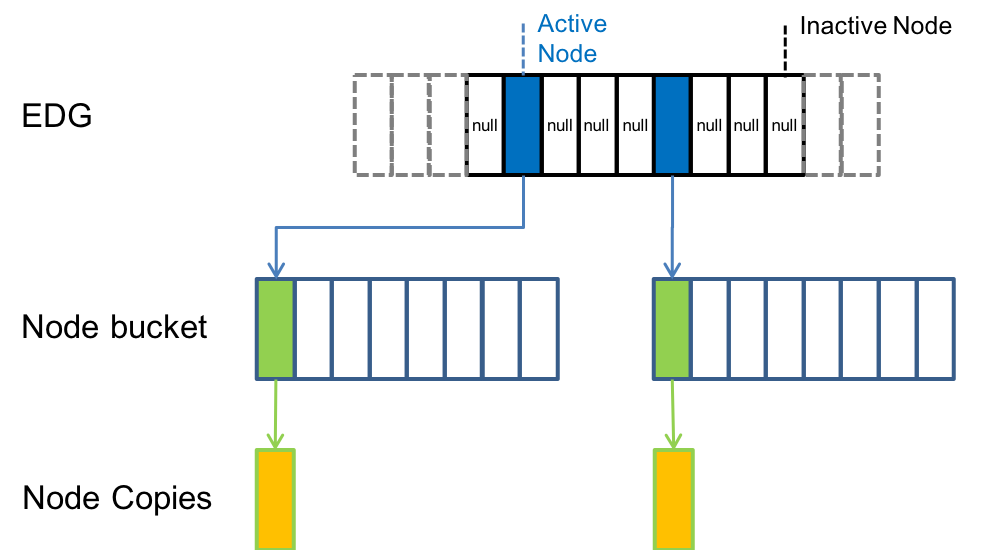}
	\caption{Illustration of re-designed EDG data structure for topological changes.}
	\label{fig:EDG_DS}
\end{figure}

To meet all these requirement, for EDG, we add a node bucket as an intermediate level shown in Figure~\ref{fig:EDG_DS}.
This node bucket has a fixed size which is the max number of node copies we allow in our system.
All EDG node buckets are stored in a flat vector. Given a 1D index $i$ of EDG node, if the pointer to a node bucket in this indexed entry is null, it means this node is inactive.
If the pointer is not null, it means at least one copy of this node is active.
The index of a node copy could be computed as $8*i+offset$.
The following is the c++ code for our new data structure:
\small
\begin{lstlisting}[language=C++, caption={C++ code using listings}]
Struct Node {
    //local translation
    Vector3f translate; 
    //local rotation
    Matrix3f rotate; 
    Node* neighbors; 
    //index1d: 1D index of the node; 
    //bucket_offset: offset of in NodeBucket.
    Int2 index {index1d,bucket_offset} 
    //offsets of 8 nodes sharing the same
    //cell with this node as the 
    //left-front-bottom one
    array<int,8> cell_offsets;
    //Real or virtual node
    bool realOrVirtual;
    bool activeOrNot;
    //sequential id in the graph
    //only used for duplicate and merge
    int parentID; 
};
Struct NodeBucket {
    Node* nodeCopies[8];
};
vector<NodeBucket *> DeformGraph;
\end{lstlisting}
\normalsize
In this way, each cell just needs to maintain the left-front-bottom node, by visiting ``cell\_offsets'' and mathematically computing the ``index1d'' of all 8 nodes based on regular grid, we could get the mapping from the cell to all its contained nodes.
The combination of ``index1d'' and ``cell\_offset'' indicates the location of a node in the data structure.

After initialization, when there is no duplicate cell, each NodeBucket only contains one node copy when this node is active.

\begin{figure*}[h]
	\centering
	\includegraphics[width=1.0\textwidth]{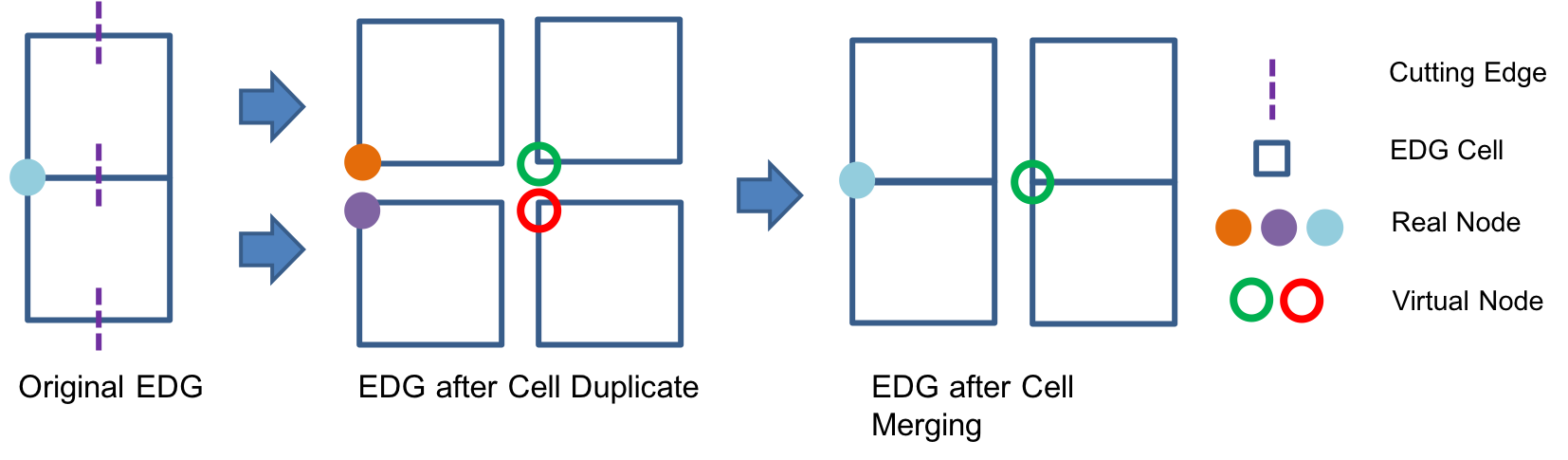}
	\caption{Steps to duplicate EDG cells and merge them.}
	\label{fig:EDG_steps}
\end{figure*}

\begin{figure*}[h]
	\centering
	\includegraphics[width=0.37\textwidth]{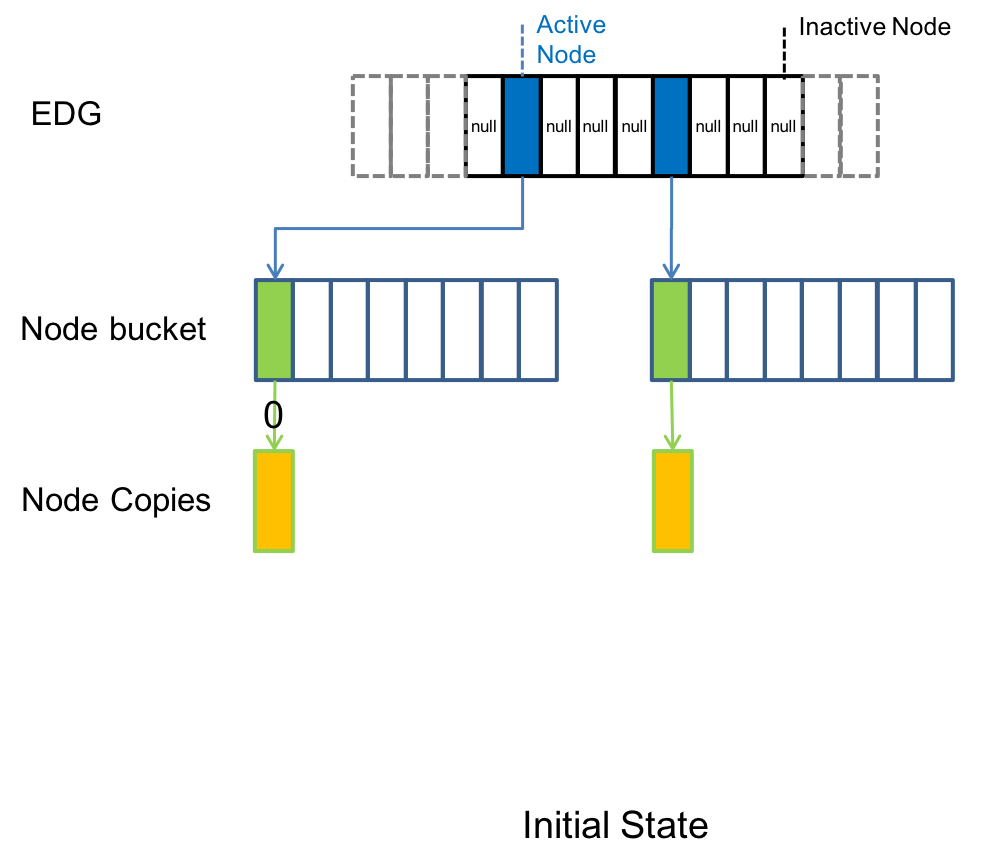}
	\includegraphics[width=0.30\textwidth]{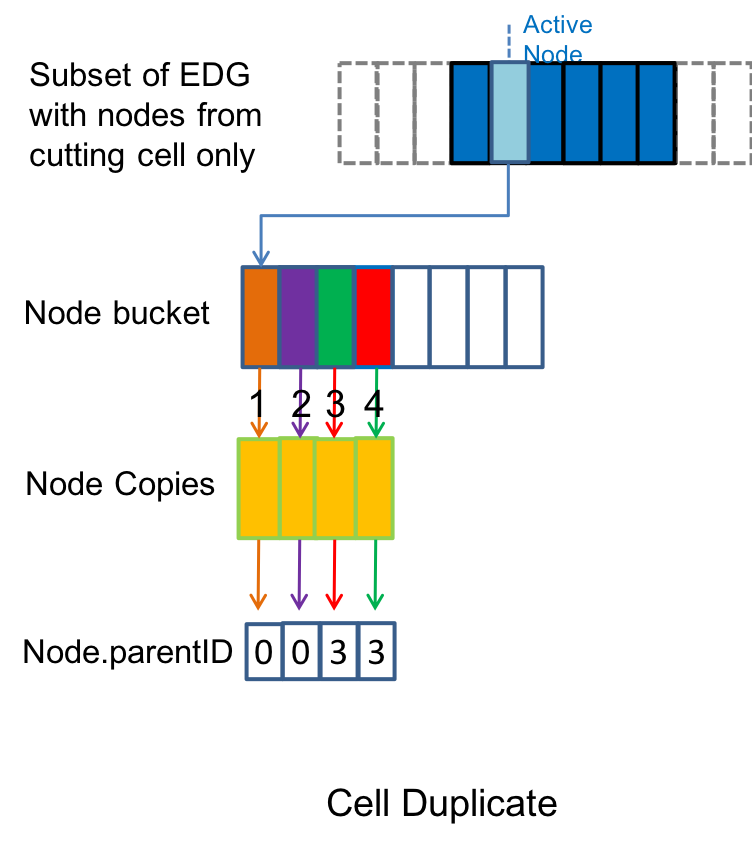}
	\includegraphics[width=0.30\textwidth]{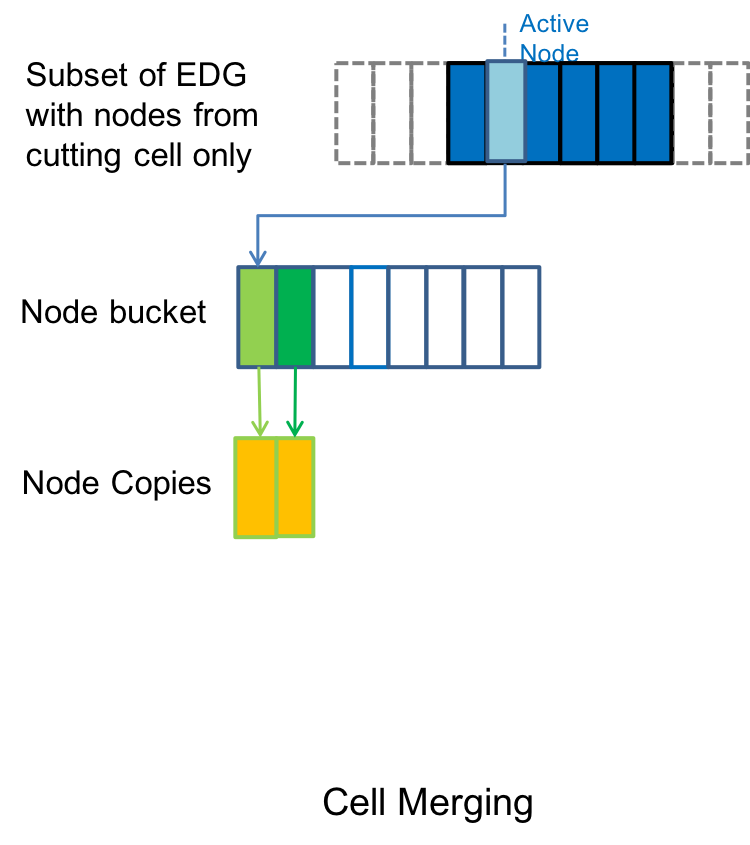}
	\caption{Illustration of steps to duplicate EDG cells and merge them by our re-designed data structure.}
	\label{fig:EDG_DS_steps}
\end{figure*}

Figure~\ref{fig:EDG_steps} and Figure~\ref{fig:EDG_DS_steps} shows the steps to duplicate and merge EDG cells.
Several strategies are used to improve the performance.
First, we only consider cells containing cutting edges, which is a small portion of the entire set of active EDG nodes.
In this step, new vector of NodeBucket will be created which only contains nodes from cutting cells.
Secondly, in the cell duplicate step, we create node copies according to the number of connected components in each cutting node cell in EDG. Shown in Figure~\ref{fig:EDG_DS_steps} ``Cell Duplicate'', the \textcolor{myLightBlue}{light blue} node is duplicate into 4 nodes: one real node (\textcolor{myOrange}{Orange}) and one virtual node (\textcolor{myPurple}{Purple}) from the top cell; one real node (\textcolor{myGreen}{Green}) and one virtual node (\textcolor{red}{Red}) from the bottom cell. 
Their parentIDs will be recorded, which are the offsets of the nodes that they inherit from.
In the case shown in Figure~\ref{fig:EDG_DS_steps}, because there is already one node copy existing in the original EDG NodeBucket vector, the offset of new node copies starting from 1.
(\textcolor{myOrange}{Orange}) node and (\textcolor{myPurple}{purple}) node are all real nodes and inherit from node 0, so their parentID is 0.
(\textcolor{myGreen}{Green}) node and (\textcolor{red}{red}) node are all virtual nodes and inherit from node 0, but they will not be merged to the parent node 0, so their parentID is 3, which is the offset of the (\textcolor{myGreen}{Green}) node.
Thirdly, in the cell merging step, we could just use UnionFind to merge all node copies of each NodeBucket individually based on their parentIDs (shown in Figure~\ref{fig:EDG_DS_steps} ``Cell Merging'').

\subsection{TSDF Volume}
We use a similar way to represent our new TSDF volume data structure.
The following is the c++ code for our new data structure:
\small
\begin{lstlisting}[language=C++, caption={C++ code using listings}]
Struct Voxel {
    float depth;
    float weight;
    Vector3i RGB; //if needed
    Vector3f warped_pos;
    Int4 index {voxel_index1d,voxel_bucket_offset,
    node_index1d, node_bucket_offset};
    array<int,8> voxel_offsets;
    bool realOrVirtual;
    //sequential id in the graph
    //only used for duplicate and merge
    int parentID;
};
Struct VoxelBucket {
    Voxel* voxelCopies[8];
};
vector<VoxelBucket *> TSDFVolume;
\end{lstlisting}
\normalsize
When we doing cell duplicate and merging, the belonged EDG cell of each voxel could be recorded.
When we doing marching cubes based mesh extraction, fast vertex/voxel to EDG cell mapping could be passed from voxel to vertex by recording the id of the left-front-bottom node in belonged EDG cell in ``Voxel.index.node\_index1d'' and ``Voxel.index.node\_bucket\_offset''.
Fast volume cell to voxel mapping is maintained in as similar way as the EDG cell to node mapping by using the property ``Voxel.voxel\_offsets''.

\clearpage
%
%